\documentclass[lettersize,journal]{IEEEtran}
\usepackage[T1]{fontenc}
\usepackage{amsmath,amsfonts,amssymb}
\usepackage{array}
\usepackage[caption=false,font=normalsize,labelfont=sf,textfont=sf]{subfig}
\usepackage{textcomp}
\usepackage{stfloats}
\usepackage{url}
\usepackage{verbatim}
\usepackage{graphicx}
\usepackage{cite}
\usepackage{booktabs}
\usepackage{amsthm}
\newtheorem{proposition}{Proposition}
\usepackage[table]{xcolor}
\usepackage[hidelinks]{hyperref} 
\usepackage{multirow} 

\begin{document}

\title{Adaptive Test-Time Memory Allocation for Long-Tailed Human--Object Interaction Detection}

\author{Yuqiu~Jiang,
        Xiaozhen~Qiao,
        Ye~Zheng,
        Yifan~Chen,
        Zhe~Sun,
        and~Xuelong~Li, \textit{Fellow, IEEE}%
\thanks{This work has been submitted to the IEEE for possible publication. Copyright may be transferred without notice, after which this version may no longer be accessible.}%
\thanks{(Corresponding authors: Zhe Sun, Xuelong Li.)}%
\thanks{Yuqiu Jiang is with the College of Future Information Technology, Fudan University, Shanghai 200433, China. This work was done during a research internship at the Institute of Artificial Intelligence, China Telecom (TeleAI), Shanghai 200232, China (E-mail: jyqiu03@gmail.com).}%
\thanks{Xiaozhen Qiao is with the School of Information Science and Technology, University of Science and Technology of China, Hefei 230026, China, and also with the Institute of Artificial Intelligence, China Telecom (TeleAI), Shanghai 200232, China.}%
\thanks{Yifan Chen is with the College of Future Information Technology, Fudan University, Shanghai 200433, China, and also with the Institute of Artificial Intelligence, China Telecom (TeleAI), Shanghai 200232, China.}%
\thanks{Ye Zheng, Zhe Sun, and Xuelong Li are with the Institute of Artificial Intelligence, China Telecom (TeleAI), Shanghai 200232, China (Corresponding authors: sunzhe@nwpu.edu.cn, xuelong\_li@ieee.org).}}%

\maketitle

\begin{abstract}
Long-tailed human--object interaction (HOI) detection remains challenging because rare interactions are sparsely represented in training data. Although test observations provide additional visual evidence, existing cache-based methods typically assign uniform memory capacity across classes, limiting their ability to capture informative features of underrepresented interactions.
To address this issue, we propose Adaptive Diversity Cache (ADC), a training-free, plug-and-play framework that adaptively constructs test-time caches through frequency-aware capacity allocation and confidence-diversity joint sample selection. Specifically, Frequency-Aware Cache Adaptation assigns larger capacities to underrepresented interactions using frequency-dependent priorities and a weighted logarithmic allocation objective, and fills remaining slots with selected augmented views. Confidence-Diversity Joint Cache Selection further combines prediction confidence with local feature diversity to retain informative cache entries. Cache retrieval then refines the predictions of the frozen detector.
Experiments on HICO-DET and V-COCO across three base detectors demonstrate consistent improvements. On HICO-DET, ADC improves mean average precision (mAP) for rare interactions by 4.02--8.57 percentage points, while also improving performance on non-rare interactions. Online adaptation further achieves gains of 2.87--7.36 percentage points on rare interactions. These results demonstrate that adaptive test-time memory effectively complements existing HOI detectors and improves long-tailed detection performance.
\end{abstract}

\begin{IEEEkeywords}
Human--object interaction detection, long-tailed recognition, test-time adaptation, memory allocation.
\end{IEEEkeywords}

\section{Introduction}
\IEEEPARstart{H}{uman--object} interaction (HOI) detection is an important step toward understanding human activities in relation to the surrounding environment~\cite{gao2018ican,liao2022gen,zou2026pose}. It supports applications in autonomous driving, robotic manipulation, and activity analysis~\cite{fang2023hodn,liu2022interactiveness,song2025embodied,sun2022human}. Recent advances in vision--language models (VLMs) have strengthened HOI recognition by transferring visual and semantic knowledge from large-scale pretraining~\cite{radford2021learning,ning2023hoiclip,lei2024ez}. Despite this progress, recognizing the full range of interactions remains difficult when training examples are distributed unevenly across verb--object compositions.

\begin{figure}[!t]
\centering
\includegraphics[width=\linewidth]{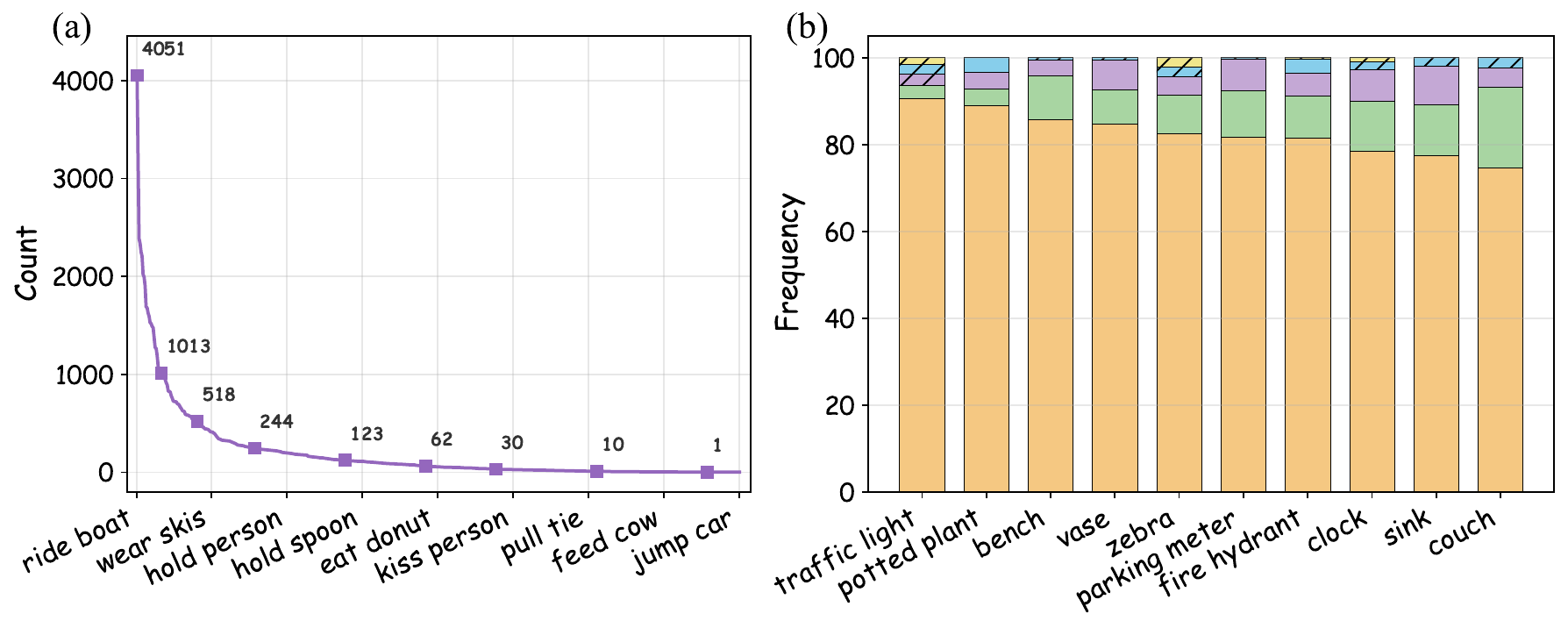}
\caption{Long-tailed interaction frequencies in HICO-DET. (a) Selected verb--object categories have substantially different training counts. (b) Verb frequencies are also imbalanced within individual object categories.}
\label{fig:long-tail}
\end{figure}

HOI datasets exhibit substantial imbalance across verb--object compositions~\cite{deng2026prototype}. As illustrated in Fig.~\ref{fig:long-tail}, some interactions occur thousands of times, whereas many others have only a few examples. The imbalance also appears among verbs associated with the same object. Moreover, frequent verbs and objects can form rare combinations, making compositional sparsity an additional source of difficulty~\cite{hou2021detecting}. With limited examples of these combinations, a detector may learn strong representations for common interactions while remaining less sensitive to rare ones. This uneven coverage limits reliable recognition across the interaction vocabulary.

\begin{figure}[!t]
\centering
\includegraphics[width=\linewidth]{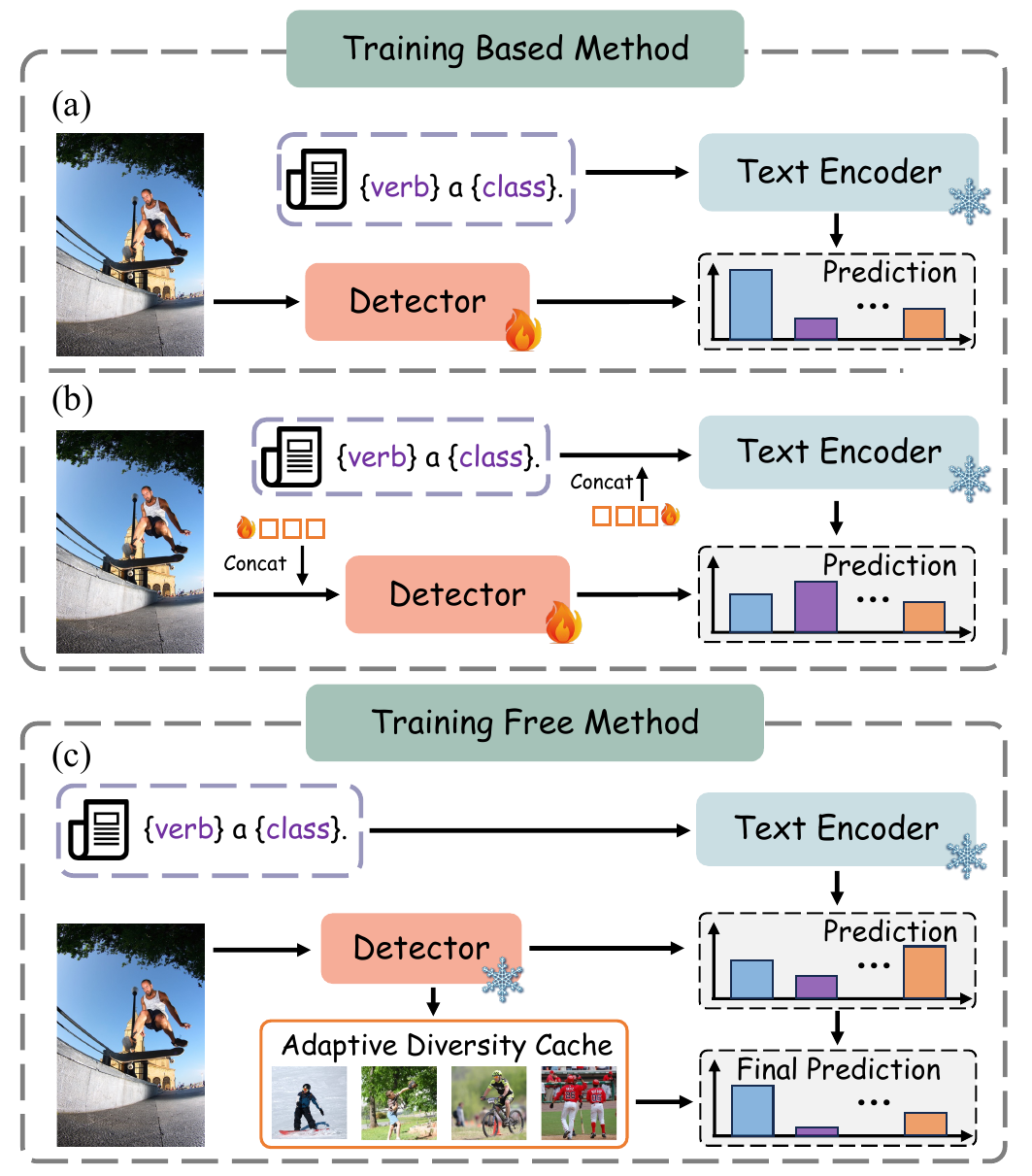}
\caption{Comparison of three HOI detection approaches. (a) Feature alignment. (b) Prompt tuning. (c) Training-free prediction support through the proposed external cache.}
\label{fig:comparison}
\end{figure}

Existing methods address this imbalance during training through compositional learning~\cite{hou2021detecting}, prior-guided decoding~\cite{gao2024dual}, and feature generation~\cite{guo2024unseen}. These strategies improve the representations learned from imbalanced data. Test observations offer a complementary source of visual evidence after training. A plug-and-play module can exploit this evidence without changing the training objective and can complement detectors that already use training-time strategies. Cache-based test-time adaptation (TTA) exploits this evidence through forward inference and external memory, without additional training. As illustrated in Fig.~\ref{fig:comparison}, it operates alongside the detector during inference and requires neither backpropagation nor parameter updates. The remaining challenge is to use the available memory effectively under a limited storage budget.

Cache-based TTA retains selected test features and uses similar observations to support subsequent predictions. TDA~\cite{karmanov2024efficient} and BoostAdapter~\cite{zhang2024boostadapter} demonstrate the value of this approach, but their standard class queues share a common capacity. In long-tailed HOI detection, however, equal capacities do not reflect the markedly different training support available to each interaction. Larger caches can provide underrepresented compositions with a broader set of visual references. Their benefit nevertheless depends on which entries are stored, since additional slots may otherwise be occupied by repeated or uncertain predictions. Effective cache construction must therefore coordinate the capacity assigned to each interaction with the observations retained within it.

We formulate test-time cache construction for long-tailed HOI detection as a constrained memory allocation problem and propose Adaptive Diversity Cache (ADC) to address it. Frequency-Aware Cache Adaptation (FACA) uses training class frequencies to assign greater capacity to underrepresented interactions and supplements unused slots with selected augmented views. Confidence-Diversity Joint Cache Selection (CJCS) retains observations with high prediction confidence and local feature variation. The formulation first distributes capacity according to unequal training support, then selects visual references within each allocation. Weighted logarithmic utility characterizes the capacity rule, and a constrained selection relaxation yields the confidence-diversity ranking. The resulting memory connects the class allocation to the observations used for prediction. ADC can be integrated with different HOI detectors through their pair features, interaction scores, and object predictions.

The main contributions are summarized as follows.
\begin{itemize}
\item We formulate test-time cache construction for long-tailed HOI detection as a constrained memory allocation problem. The formulation links class capacities and retained observations through a shared storage budget, providing the basis for the training-free, plug-and-play ADC framework.
\item We develop FACA and CJCS for capacity allocation and sample selection, respectively. FACA prioritizes classes with limited training support through a bounded capacity rule, while CJCS ranks candidates by confidence and local feature variation. Their rules are characterized through weighted logarithmic utility and constrained selection.
\item Experiments on HICO-DET and V-COCO demonstrate consistent improvements across three base detectors, with particularly strong gains on rare interactions. Improvements on unseen compositions and in online adaptation further establish the applicability of the framework.
\end{itemize}

\section{Related Work}
\subsection{Human--Object Interaction Detection}
HOI detection has advanced through improved human--object association, transferable visual representations, and compositional learning. Transformer-based detectors jointly model pair associations and interaction context~\cite{kim2021hotr,tamura2021qpic,zhang2022efficient,zhang2023exploring,jia2025contexthoi,yang2025sibling}. VLMs further extend the semantic knowledge available for recognition beyond the annotated interaction examples. GEN-VLKT~\cite{liao2022gen} transfers visual--language knowledge through distillation, HOICLIP~\cite{ning2023hoiclip} exploits pretrained visual and semantic representations, and CLIP4HOI~\cite{mao2023clip4hoi} adapts CLIP for zero-shot HOI detection. Prompt learning offers another means of adapting pretrained representations, including the guided prompts used in EZ-HOI~\cite{lei2024ez}.

Uneven coverage of verb--object compositions has motivated learning strategies specifically designed for rare and unseen interactions~\cite{zhang2024cefa}. Fabricated compositional learning~\cite{hou2021detecting} constructs additional compositions to expand the training space. Dual-prior augmented decoding~\cite{gao2024dual} incorporates prior information for long-tail recognition. HOIGEN~\cite{guo2024unseen} uses CLIP to generate features and jointly trains on real and generated features to improve recognition of seen and unseen interactions. These approaches improve the representations learned from imbalanced data, although the scarcity of examples for individual compositions remains a challenge. Their training objectives primarily determine how interaction knowledge is learned before inference.

Stored visual references have also been used for HOI recognition. ADA-CM~\cite{lei2023efficient} constructs concept-guided memory from training samples and provides both training-free and lightweight trainable modes. This reference memory makes interaction knowledge from the training data available for retrieval during inference. Despite progress in representation learning and memory-based recognition, compositions with limited or no training examples remain difficult to recognize.

\subsection{Test-Time Adaptation}
Test-time adaptation uses unlabeled observations to adjust a trained model or its predictions to the data encountered during inference. Early work emphasizes parameter updates through entropy minimization~\cite{wang2021tent}, auxiliary self-supervision~\cite{sun2020test}, or contrastive learning~\cite{chen2022contrastive}. Related methods have been developed for semantic segmentation~\cite{hu2021fully,shin2022mm}, super-resolution~\cite{deng2023efficient,shocher2018zero}, and object detection~\cite{ruan2024fully}. Parameter optimization requires backward computation and an adaptation objective suited to the prediction task. This motivates training-free alternatives that use test data while retaining the learned model parameters.

For VLMs, TDA~\cite{karmanov2024efficient} maintains positive and negative key--value caches with uncertainty-based updates. BoostAdapter~\cite{zhang2024boostadapter} combines historical memory with augmented regional views of the current sample. Both use feature retrieval to incorporate information from test observations into the prediction. Their standard queue configurations impose the same shot limit on each class, providing a simple storage constraint across the label space.

For long-tailed HOI detection, however, a shared limit does not reflect the unequal training support of different verb--object compositions. Moreover, confidence-based selection can retain near-duplicate features, reducing the variety represented within a small queue. Augmented views expand the candidate pool, but their usefulness still depends on the available class capacity and the selection criterion. Managing these interdependent decisions under a bounded memory budget remains an open issue when transferring cache-based adaptation to long-tailed HOI detection.

\section{Methodology}
\begin{figure*}[!t]
\centering
\includegraphics[width=\linewidth]{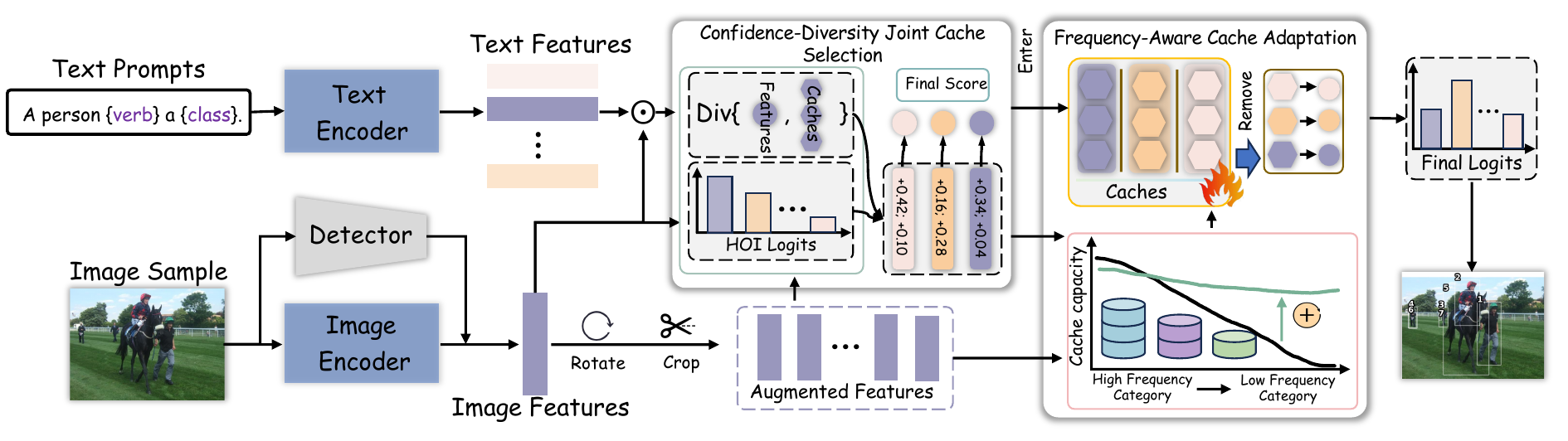}
\caption{Overview of ADC. A frozen HOI detector provides pair features and interaction predictions. FACA assigns capacities using training class frequencies and supplements underfilled caches with selected augmented views. CJCS retains candidates according to confidence and local feature diversity. The stored features and class assignments support affinity-based retrieval, and the retrieved class scores are combined with the base predictions.}
\label{fig:pipeline}
\end{figure*}

We propose Adaptive Diversity Cache (ADC) for test-time memory allocation in long-tailed HOI detection. As illustrated in Fig.~\ref{fig:pipeline}, ADC consists of Frequency-Aware Cache Adaptation (FACA) and Confidence-Diversity Joint Cache Selection (CJCS). FACA assigns class capacities from training frequencies and supplements underfilled caches with selected augmented views. Within these capacities, CJCS retains entries according to prediction confidence and local feature diversity. Affinity-based retrieval then combines the stored features and class assignments to refine the base predictions. All operations use the outputs of the frozen detector.

\subsection{Problem Formulation}
\label{sec:formulation}
With the detector fixed, test-time adaptation acts on the visual evidence retained for prediction. Under a limited memory budget, the training imbalance creates a resource-allocation problem. Classes with few training examples receive limited supervision, while the benefit of additional cache space depends on the observations retained. Cache construction must therefore determine both the capacity reserved for each interaction and the features retained within that capacity. These decisions are linked by the same storage constraint.

To organize the observations by interaction class, a representative label is assigned to each detected human--object pair according to
\begin{equation}
\ell_i=\arg\max_c p_i^{(c)},
\label{eq:assignment}
\end{equation}
where $c=(o,v)$ denotes a valid object--verb composition. The probability vector $p_i\in[0,1]^C$ is obtained by applying softmax to the current interaction logits $h_i$ over classes compatible with the predicted object, with zeros assigned elsewhere. This distribution is used for cache assignment and entropy computation. Each candidate entry contains the pair feature $f_i\in\mathbb R^d$ and interaction logits $h_i\in\mathbb R^C$. The representative label indexes its class cache, while the detector can still predict several interactions for the same pair.

At a memory update, the available entries form a candidate pool for each class. Cache construction then selects class capacities and retained subsets subject to
\begin{equation}
\begin{aligned}
&\mathcal Q_c\subseteq\mathcal U_c,\qquad |\mathcal Q_c|\leq K_c,\\
&K_{\min}\leq K_c\leq K_{\max},\qquad
\sum_{c=1}^{C}K_c\leq B,
\end{aligned}
\label{eq:budget}
\end{equation}
where $\mathcal U_c$ denotes the current candidate pool, $\mathcal Q_c$ denotes the retained subset, $K_c$ is an integer capacity, and $B$ is the total slot limit. The bounds satisfy $1\leq K_{\min}\leq K_{\max}$. The capacity $K_c$ specifies the reserved space, whereas the occupancy $|\mathcal Q_c|$ is determined by the available entries. Reserving capacity allows classes with few current observations to accommodate subsequent updates and selected augmented views.

The allocation is guided by training support and diminishing returns from additional slots. A positive class priority is combined with a logarithmic utility to give the continuous allocation problem
\begin{equation}
\begin{aligned}
\max_{\boldsymbol k}\quad &U(\boldsymbol k)=\sum_c r_c\log k_c,\\
\text{subject to}\quad &\sum_c k_c\leq B,\quad
K_{\min}\leq k_c\leq K_{\max},
\end{aligned}
\label{eq:utility}
\end{equation}
where $k_c$ is a continuous relaxation of capacity and $r_c>0$ denotes a priority determined by the training frequency of class $c$. The marginal utility $r_c/k_c$ is larger for a higher-priority class and decreases as its allocation grows. This provides a basis for distributing finite memory across classes with unequal training support.

Once the class capacities are assigned, the content decision maximizes the value of the retained entries within each allocation,
\begin{equation}
\begin{aligned}
\max_{\mathcal S\subseteq\mathcal U_c}\quad
&J_c(\mathcal S)=\sum_{i\in\mathcal S}u_i,\\
\text{subject to}\quad &|\mathcal S|=m_c,\qquad
m_c=\min\{K_c,|\mathcal U_c|\},
\end{aligned}
\label{eq:content_objective}
\end{equation}
where $u_i$ measures prediction confidence and local feature diversity on the current pool. Confidence favors predictions with lower uncertainty, while local variation distinguishes candidates that offer different visual references. The two objectives are addressed in two stages: FACA converts frequency priorities into bounded integer capacities, and CJCS selects entries within these capacities using the joint score. Cache completion supplies additional candidates when reserved capacity remains unfilled.

\subsection{Frequency-Aware Cache Adaptation}
FACA implements the allocation objective in Eq.~\eqref{eq:utility} through class priorities derived from training frequencies. The resulting capacities reserve more space for underrepresented interactions while maintaining lower and upper limits for every class. Selected augmented views supplement caches with fewer observations than their assigned capacities.

To assign a larger priority to classes with limited training support, a frequency-dependent weight is defined as
\begin{equation}
r_c=\left(\frac{n_{\max}+\delta}{n_c+\delta}\right)^{\eta}
\exp\left(-\lambda\frac{n_c}{N}\right),
\label{eq:priority}
\end{equation}
where $n_c$ denotes the number of training instances of class $c$, $n_{\max}=\max_c n_c$, and $N=\sum_c n_c>0$. The smoothing constant $\delta>0$ prevents a zero denominator. The exponent $\eta\in(0,1]$ controls the strength of inverse-frequency weighting, while $\lambda\geq0$ adjusts the relative-frequency penalty. These class statistics remain fixed throughout test-time adaptation.

The priority ordering follows directly from the relative weights of two classes,
\begin{equation}
\frac{r_a}{r_b}
=\left(\frac{n_b+\delta}{n_a+\delta}\right)^\eta
\exp\left(\lambda\frac{n_b-n_a}{N}\right).
\label{eq:ratio}
\end{equation}
For $n_a<n_b$, the ratio exceeds one, giving $r_a>r_b$. The inverse-frequency factor prioritizes underrepresented classes, and its exponent controls sensitivity to frequency differences.

Using the frequency-dependent weights $r_c$ defined in Eq.~\eqref{eq:priority}, the allocation problem in Eq.~\eqref{eq:utility} admits the following optimal continuous solution.

\begin{proposition}[Continuous capacity allocation]
For positive weights and $CK_{\min}<B<CK_{\max}$, the unique optimal capacity vector of Eq.~\eqref{eq:utility} is
\begin{equation}
k_c^*=\operatorname{Clip}\left(\frac{r_c}{\mu},K_{\min},K_{\max}\right),
\label{eq:waterfill}
\end{equation}
where $\mu>0$ is chosen to satisfy $\sum_c k_c^*=B$.
\end{proposition}
The result follows from strict concavity and the Karush--Kuhn--Tucker conditions, as proved in Appendix~\ref{app:allocation}. Interior capacities are proportional to priority. The lower bound preserves a minimum allocation for every class, and the upper bound limits the concentration of memory. Because clipping and rounding are monotone, the integer rule below also preserves the frequency ordering, giving $K_a\geq K_b$ whenever $n_a<n_b$.

The integer capacity rule is obtained by scaling, clipping, and rounding,
\begin{equation}
K_c=\operatorname{Clip}\left(\lfloor K_{\mathrm{base}}r_c\rfloor,
K_{\min},K_{\max}\right),
\label{eq:capacity}
\end{equation}
where $K_{\mathrm{base}}>0$ controls the overall scale. For a prescribed budget, $K_{\mathrm{base}}=1/\mu$ recovers Eq.~\eqref{eq:waterfill} before rounding. For a fixed scale, the continuous allocation corresponds to the induced budget $B_0=\sum_c\operatorname{Clip}(K_{\mathrm{base}}r_c,K_{\min},K_{\max})$. Thus, the same capacity family can be interpreted through either a memory budget or a fixed allocation scale. Appendix~\ref{app:allocation} gives the associated rounding bound.

When a class has fewer observations than its assigned capacity, augmented views are generated from the entries already in its cache. Low-entropy views form an eligible candidate set and are ranked using the confidence-diversity score introduced in Sec.~\ref{sec:cjcs}. The completion rule is
\begin{equation}
\begin{aligned}
\mathcal Q_c^+&=\mathcal Q_c\cup\operatorname{TopK}_{u}(\mathcal A_c,a_c),\\
a_c&=\min\{K_c-|\mathcal Q_c|,|\mathcal A_c|\},
\end{aligned}
\label{eq:completion}
\end{equation}
where $\mathcal A_c$ contains eligible augmented views whose own predicted representative label is $c$, $u$ denotes their joint selection score computed on $\mathcal Q_c\cup\mathcal A_c$, and $a_c$ is the number admitted. After completion, $\mathcal Q_c$ is replaced by $\mathcal Q_c^+$, and the admitted augmented entries participate in subsequent cache updates.

\subsection{Confidence-Diversity Joint Cache Selection}
\label{sec:cjcs}
CJCS instantiates the content objective in Eq.~\eqref{eq:content_objective} within the capacities assigned by FACA. Its score combines prediction confidence with local feature diversity. Among candidates with similar confidence, the diversity term favors features with different directions. For an incoming entry assigned to class $c$, the temporary pool is formed as
\begin{equation}
\mathcal U_c=\mathcal Q_c\cup\{(f_i,h_i)\},\qquad M_c=|\mathcal U_c|,
\label{eq:pool}
\end{equation}
where $M_c$ denotes the pool size. Confidence and diversity are evaluated on this pool before the retained entries are updated.

Prediction certainty is measured using normalized entropy,
\begin{equation}
\begin{aligned}
H(p_i)&=-\sum_{a=1}^{C}p_i^{(a)}\log p_i^{(a)},\\
s_i&=1-\frac{H(p_i)}{H_c^{\max}},\qquad
H_c^{\max}=\max_{j\in\mathcal U_c}H(p_j),
\end{aligned}
\label{eq:confidence}
\end{equation}
where $s_i=\mathcal S_{\mathrm{conf}}(p_i)\in[0,1]$ denotes the confidence score and $0\log0=0$. If every entropy is zero, $s_i$ is set to one. Lower-entropy predictions receive higher scores, with normalization performed within the current class pool.

To retain varied observations without favoring feature distance alone, angular dissimilarity is weighted by local proximity,
\begin{equation}
\begin{aligned}
d_i=\frac{1}{M_c}\sum_{\substack{j\in\mathcal U_c\\j\ne i}}
&\left(1-\frac{f_i^\top f_j}{\|f_i\|_2\|f_j\|_2}\right)\\[-1mm]
&\times\exp\left(-\gamma\|f_i-f_j\|_2^2\right),
\end{aligned}
\label{eq:diversity}
\end{equation}
where $d_i=\mathcal S_{\mathrm{div}}(f_i)$ denotes the diversity score, $\gamma>0$ controls locality, and features are assumed to be nonzero. Nearly aligned features make only a small contribution, while the proximity kernel attenuates distant pairs. The score therefore emphasizes variation among locally related observations. The relationship between the two factors is derived in Appendix~\ref{app:selection}.

To determine the entry score in Eq.~\eqref{eq:content_objective}, local diversity is maximized subject to a confidence requirement,
\begin{equation}
\begin{aligned}
\max_{\mathcal S\subseteq\mathcal U_c}\quad&D(\mathcal S)=\sum_{i\in\mathcal S}d_i,\\
\text{subject to}\quad&Q(\mathcal S)=\sum_{i\in\mathcal S}s_i\geq q_0,\quad
|\mathcal S|=m_c,
\end{aligned}
\label{eq:selection_constraint}
\end{equation}
where $m_c=\min\{K_c,M_c\}$ is the number of retained entries and $q_0$ specifies the confidence requirement. Introducing a multiplier $\nu\geq0$ yields the relaxed objective $D(\mathcal S)+\nu Q(\mathcal S)$ up to a constant. Dividing by $1+\nu$ gives the joint ranking score
\begin{equation}
u_i=\tau d_i+(1-\tau)s_i,\qquad \tau=\frac{1}{1+\nu},
\label{eq:joint}
\end{equation}
where $\tau\in(0,1]$ controls the relative contribution of diversity. Confidence-only selection is obtained at the limiting value $\tau=0$ as $\nu\to\infty$. For a fixed candidate pool, the relaxed objective gives the entry scores used in Eq.~\eqref{eq:content_objective}.

\begin{proposition}[Selection for a fixed candidate pool]
\label{prop:selection}
With $d_i$ and $s_i$ fixed on $\mathcal U_c$, the update
\begin{equation}
\mathcal Q_c\leftarrow\operatorname{TopK}_{u}(\mathcal U_c,m_c)
\label{eq:update}
\end{equation}
maximizes $\sum_{i\in\mathcal S}u_i$ over all subsets of size $m_c$. For $0<\tau<1$, the selected set is Pareto optimal for the objectives $(D,Q)$ on that pool.
\end{proposition}
The proof is given in Appendix~\ref{app:selection}. An exchange argument establishes optimality for the weighted score, and positive weights ensure that an improvement in either objective without reducing the other would improve the joint objective. Scores are recomputed when the pool changes, so selection follows the current cache contents.

The selected entries support prediction through feature retrieval. For a query pair with predicted object $o$, the compatible memory is formed by the union $\mathcal S_o=\bigcup_{c=(o,v)}\mathcal Q_c^+$. Its keys and values are
\begin{equation}
\mathbf K=[\hat f_1^\top;\ldots;\hat f_{M_o}^\top],\qquad
\mathbf V=[e_{\ell_1}^\top;\ldots;e_{\ell_{M_o}}^\top],
\label{eq:keys}
\end{equation}
where $\hat f_i=f_i/\|f_i\|_2$ denotes a normalized feature, $M_o=|\mathcal S_o|$, and $e_{\ell_i}\in\mathbb R^C$ denotes the one-hot vector of the representative class. Following cache-based prediction~\cite{karmanov2024efficient,zhang2024boostadapter}, the retrieved support is added to the base scores according to
\begin{equation}
\begin{aligned}
w_i(q)&=\exp\{-\beta(1-q^\top\hat f_i)\},\\
z(q)&=z_{\mathrm{base}}(q)+\alpha\sum_{i\in\mathcal S_o}w_i(q)e_{\ell_i},
\end{aligned}
\label{eq:readout}
\end{equation}
where $q$ is the unit-normalized query feature, $\alpha\geq0$ scales the cache contribution, and $\beta>0$ controls affinity decay. Entries that are more similar to the query provide stronger support for their associated classes.

The contribution of a class can be separated into its occupancy and average affinity,
\begin{equation}
z_{\mathrm{cache},c}(q)=\alpha|\mathcal Q_c^+|\bar w_c(q),\qquad
0\leq z_{\mathrm{cache},c}(q)\leq\alpha K_c,
\label{eq:mass}
\end{equation}
where $\bar w_c(q)$ denotes the mean retrieval weight of its entries and an empty cache contributes zero. FACA controls the number of available slots and their occupancy through completion, while CJCS determines the features that contribute to the mean affinity. The same budget also bounds the total retrieved contribution by $\sum_c z_{\mathrm{cache},c}(q)\leq\alpha B$. Thus, the two memory decisions control both the amount of retained evidence and its relevance to the query. Retrieval sensitivity and computational cost are analyzed in Appendices~\ref{app:stability} and~\ref{app:complexity}.

\section{Experiments}
\subsection{Datasets and Metrics}
Experiments are conducted on HICO-DET~\cite{chao2018learning} and V-COCO~\cite{gupta2015visual}. HICO-DET contains 38,118 training images and 9,658 test images, with 600 interaction categories formed from 80 object categories and 117 actions. V-COCO contains 5,400 training and validation images and 4,946 test images, with 29 action categories. Together, the two benchmarks evaluate HOI recognition under different interaction vocabularies and annotation settings.

For HICO-DET, mean average precision (mAP) is reported over all 600 classes (Full), the 138 classes with fewer than ten training instances (Rare), and the remaining 462 classes (Non-rare). A detection is correct when its interaction label matches the annotation and both human and object boxes meet the intersection-over-union threshold of 0.5. For V-COCO, role average precision is reported under Scenario 1. Performance differences are expressed in percentage points.

\subsection{Implementation Details}
\label{sec:protocol}
ADC is integrated with HOIGEN~\cite{guo2024unseen}, EZ-HOI~\cite{lei2024ez}, and ADA-CM~\cite{lei2023efficient}. ADA-CM is used for component and parameter studies. The evaluated configurations use DETR~\cite{carion2020end} with a ResNet-50 backbone for object detection. Pair features, interaction scores, and predicted object categories are supplied to ADC, while all detector parameters remain frozen. Experiments are run on a single Tesla A100 GPU with 40 GB of memory.

Class frequencies for FACA are obtained from the training split used by each detector, with $n_{\max}$ and $N$ computed from the same statistics. For UC and SG evaluations, the corresponding retained training splits are used. Cache contents are updated from the features and predictions produced on unlabeled test images.

The main results use a two-pass protocol. The complete unlabeled test set is first processed to initialize the feature caches. Final predictions are produced in the second pass, during which the caches continue to be updated using the refined predictions. For online evaluation, the cache starts empty and is built incrementally during a single pass through the test stream, with retrieval preceding each cache update. The frequency prior is the same fixed training statistic in both settings.

The cache settings are $\alpha=3$, $\beta=5$, and $K_{\max}=6$. The reference configuration uses a fixed allocation scale, with
$K_{\min}=1$, $K_{\mathrm{base}}=1$, $\eta=0.4$,
$\delta=1$, and $\lambda=1$. The joint selection weight is $\tau=0.7$. The pair features used for diversity scoring and retrieval are L2-normalized, with $\gamma=1$. Cache completion generates 64 augmented candidates per sample using crop, rotation, shear, and translation, with augmentation severity set to 1. The lowest-entropy 10\% are retained as eligible candidates for joint-score selection. The same settings are used across the main detector comparisons.

\begin{table}[!t]
\centering
\caption{Performance comparison on HICO-DET and V-COCO. Paired rows report the base detector and its cache-based extensions.}
\label{tab:performance}
\small
\setlength{\tabcolsep}{2.8pt}
\resizebox{\linewidth}{!}{%
\begin{tabular}{lccccc} 
\toprule
\multirow{2}{*}{Method} & \multirow{2}{*}{Backbone} & \multicolumn{3}{c}{HICO-DET} & V-COCO \\
\cmidrule(lr){3-5} \cmidrule(lr){6-6}
& & Full & Rare & Non-rare & $AP_{role}^{\#1}$ \\
\midrule
iCAN~\cite{gao2018ican} & R50 & 14.84 & 10.45 & 16.15 & 45.3 \\
DRG~\cite{gao2020drg} & R50-F & 24.53 & 19.47 & 26.04 & 51.0 \\
HOTR~\cite{kim2021hotr} & R50 & 25.10 & 17.34 & 27.42 & 55.2 \\
QPIC~\cite{tamura2021qpic} & R50 & 29.07 & 21.85 & 31.23 & 58.8 \\
CPC~\cite{park2022consistency} & R50 & 29.63 & 23.14 & 31.57 & 63.1 \\
FGAHOI~\cite{ma2023fgahoi} & Swin-T & 29.94 & 22.24 & 32.24 & 60.5 \\
RLIP~\cite{yuan2022rlip} & R50 & 32.84 & 26.85 & 34.63 & 61.9 \\
UPT~\cite{zhang2022efficient} & R50 & 31.66 & 25.94 & 33.36 & 59.0 \\
GEN-VLKT~\cite{liao2022gen} & R50 & 33.75 & 29.25 & 35.10 & 62.4 \\
PViC~\cite{zhang2023exploring} & R50 & 34.69 & 32.14 & 35.45 & 62.8 \\
HOICLIP~\cite{ning2023hoiclip} & R50 & 34.69 & 31.12 & 35.74 & 63.5 \\
RLIPv2~\cite{yuan2023rlipv2} & R50 & 35.38 & 29.61 & 37.10 & 65.9 \\
LOGICHOI~\cite{li2023neural} & R50 & 35.47 & 32.03 & 36.22 & 64.4 \\
ViPLO~\cite{park2023viplo} & ViT-B/16 & 37.22 & 35.45 & 37.75 & 62.2 \\
SCTC~\cite{jiang2024exploring} & R50 & 37.92 & 34.78 & 38.86 & \textbf{67.1} \\
\midrule
HOIGEN~\cite{guo2024unseen} & R50 & 33.56 & 32.10 & 34.00 & - \\
+ BoostAdapter~\cite{zhang2024boostadapter} & R50 & 34.83 & 35.37 & 34.67 & - \\
\rowcolor{gray!20}
+ ADC & R50 & 37.95 & 40.67 & 37.14 & - \\
$\Delta$ & - & \textcolor{red}{+4.39} & \textcolor{red}{+8.57} & \textcolor{red}{+3.14} & - \\
\midrule
EZ-HOI~\cite{lei2024ez} & R50 & 38.33 & 36.70 & 38.82 & 60.5 \\
+ BoostAdapter~\cite{zhang2024boostadapter} & R50 & 37.48 & 34.58 & 38.35 & 61.3 \\
\rowcolor{gray!20}
+ ADC & R50 & 39.80 & \textbf{41.85} & 39.20 & 64.5 \\
$\Delta$ & - & \textcolor{red}{+1.47} & \textcolor{red}{+5.15} & \textcolor{red}{+0.38} & \textcolor{red}{+4.0} \\
\midrule
ADA-CM~\cite{lei2023efficient} & R50 & 38.39 & 37.46 & 38.67 & 58.5 \\
+ BoostAdapter~\cite{zhang2024boostadapter} & R50 & 38.75 & 39.01 & 38.68 & 59.9 \\
\rowcolor{gray!20}
+ ADC & R50 & \textbf{39.81} & 41.48 & \textbf{39.30} & 62.9 \\
$\Delta$ & - & \textcolor{red}{+1.42} & \textcolor{red}{+4.02} & \textcolor{red}{+0.63} & \textcolor{red}{+4.4} \\
\bottomrule
\end{tabular}}
\end{table}

\subsection{Comparison with Existing Methods}
Results on HICO-DET and V-COCO are reported in Table~\ref{tab:performance}. On HICO-DET, ADC improves Full mAP from 33.56 to 37.95 for HOIGEN, from 38.33 to 39.80 for EZ-HOI, and from 38.39 to 39.81 for ADA-CM. The largest gain is obtained with HOIGEN, which has the lowest initial Full mAP. EZ-HOI and ADA-CM also improve despite their stronger baseline performance, demonstrating that ADC benefits all three base detectors.

The improvements are particularly pronounced for rare interactions. Rare mAP increases by 8.57, 5.15, and 4.02 points for HOIGEN, EZ-HOI, and ADA-CM, respectively. The corresponding Non-rare gains are 3.14, 0.38, and 0.63 points. For ADA-CM, Rare mAP changes from 1.21 points below Non-rare mAP to 2.18 points above it after adaptation. EZ-HOI exhibits a similar reversal. The larger Rare gains show that ADC benefits underrepresented interactions while also improving frequent categories.

ADC achieves higher Rare mAP than BoostAdapter on all three base detectors, with margins of 5.30, 7.27, and 2.47 points for HOIGEN, EZ-HOI, and ADA-CM, respectively. On EZ-HOI, BoostAdapter reduces Rare mAP from 36.70 to 34.58, whereas ADC raises it to 41.85. These results highlight the importance of adapting cache construction to imbalanced interaction classes.

On V-COCO, ADC raises role mAP from 58.5 to 62.9 for ADA-CM and from 60.5 to 64.5 for EZ-HOI. The adapted scores exceed the corresponding BoostAdapter results by 3.0 and 3.2 points. The gains on V-COCO extend the improvements beyond the fine-grained interaction categories of HICO-DET.

\subsection{Ablation Studies}
\begin{table}[!t]
\centering
\caption{Component ablation on HICO-DET with ADA-CM. Confidence, diversity, frequency-aware capacity, and cache completion are incorporated cumulatively.}
\label{tab:ablation}
\small
\setlength{\tabcolsep}{4pt} 
\begin{tabular}{ccccccc} 
\toprule
\multicolumn{2}{c}{CJCS} & \multicolumn{2}{c}{FACA} & \multirow{2}{*}{Full} & \multirow{2}{*}{Rare} & \multirow{2}{*}{Non-rare} \\  
\cmidrule(lr){1-2} \cmidrule(lr){3-4}
\(\mathcal{S}_{\text{conf}}\) & \(\mathcal{S}_{\text{div}}\) & $K_c$ & Completion & & & \\
\midrule
\checkmark & & & & 38.49 & 38.42 & 38.51 \\
\checkmark & \checkmark & & & 38.79 & 39.63 & 38.54 \\
\checkmark & \checkmark & \checkmark & & 39.47 & 40.32 & 39.21 \\
\checkmark & \checkmark & \checkmark & \checkmark & \textbf{39.81} & \textbf{41.48} & \textbf{39.30} \\
\bottomrule
\end{tabular}
\end{table}

\subsubsection{Component Contributions}
The cumulative component results are reported in Table~\ref{tab:ablation}. Adding diversity to confidence-based selection increases Full mAP from 38.49 to 38.79 and Rare mAP from 38.42 to 39.63, yielding gains of 0.30 and 1.21 points, respectively. Non-rare performance remains largely unchanged. The larger gain on Rare classes indicates that diversity improves the selection of visual references for underrepresented interactions.

Adding frequency-aware capacity then raises Full, Rare, and Non-rare mAP by 0.68, 0.69, and 0.67 points, respectively. Both groups benefit when sample selection is coupled with class-dependent allocation. Cache completion provides a further 0.34-point Full gain, including 1.16 points on Rare and 0.09 points on Non-rare. The stronger Rare response is consistent with the use of augmentation to fill slots reserved for underrepresented interactions. The complete configuration achieves the highest result in all three columns, reaching 39.81 Full mAP and 41.48 Rare mAP.

\begin{figure}[!t]
\centering
\includegraphics[width=\linewidth]{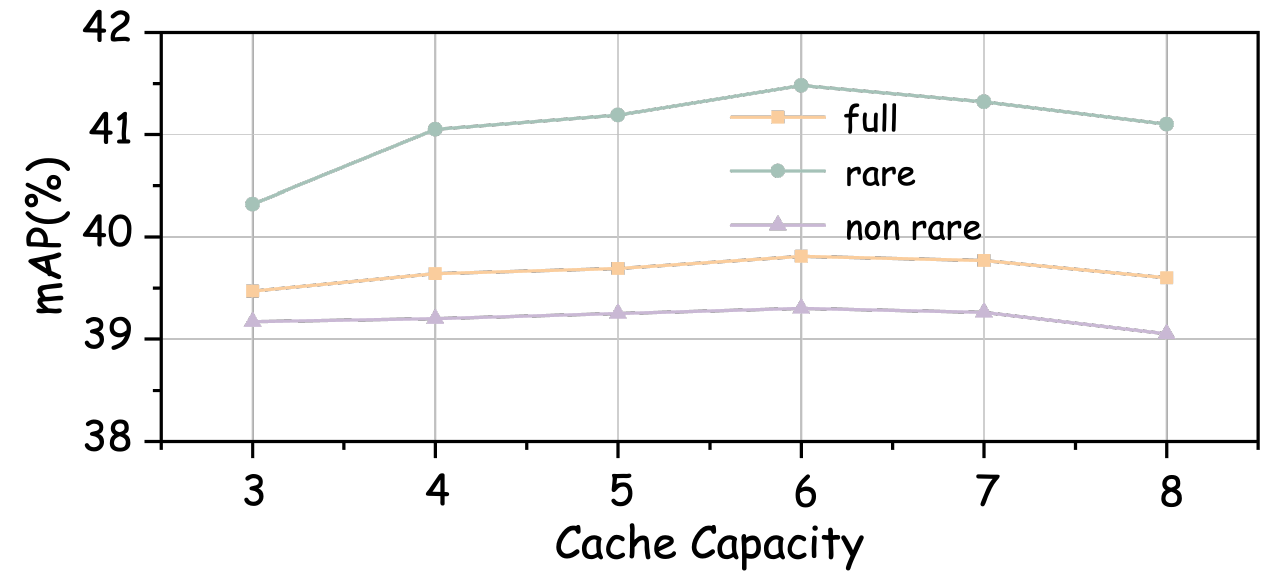}
\caption{Effect of the upper capacity bound $K_{\max}$ on detection performance. Full and Rare mAP peak at $K_{\max}=6$.}
\label{fig:capacity}
\end{figure}

\begin{figure}[!t]
\centering
\includegraphics[width=\linewidth]{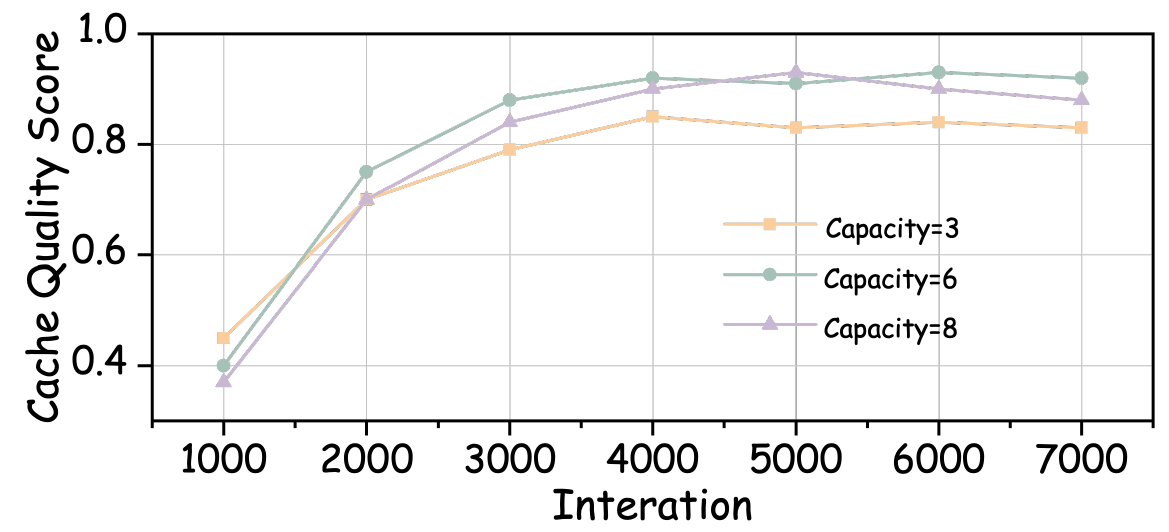}
\caption{Evolution of the normalized joint selection score during cache updates under different capacity settings.}
\label{fig:quality}
\end{figure}

\begin{figure}[!t]
\centering
\includegraphics[width=\linewidth]{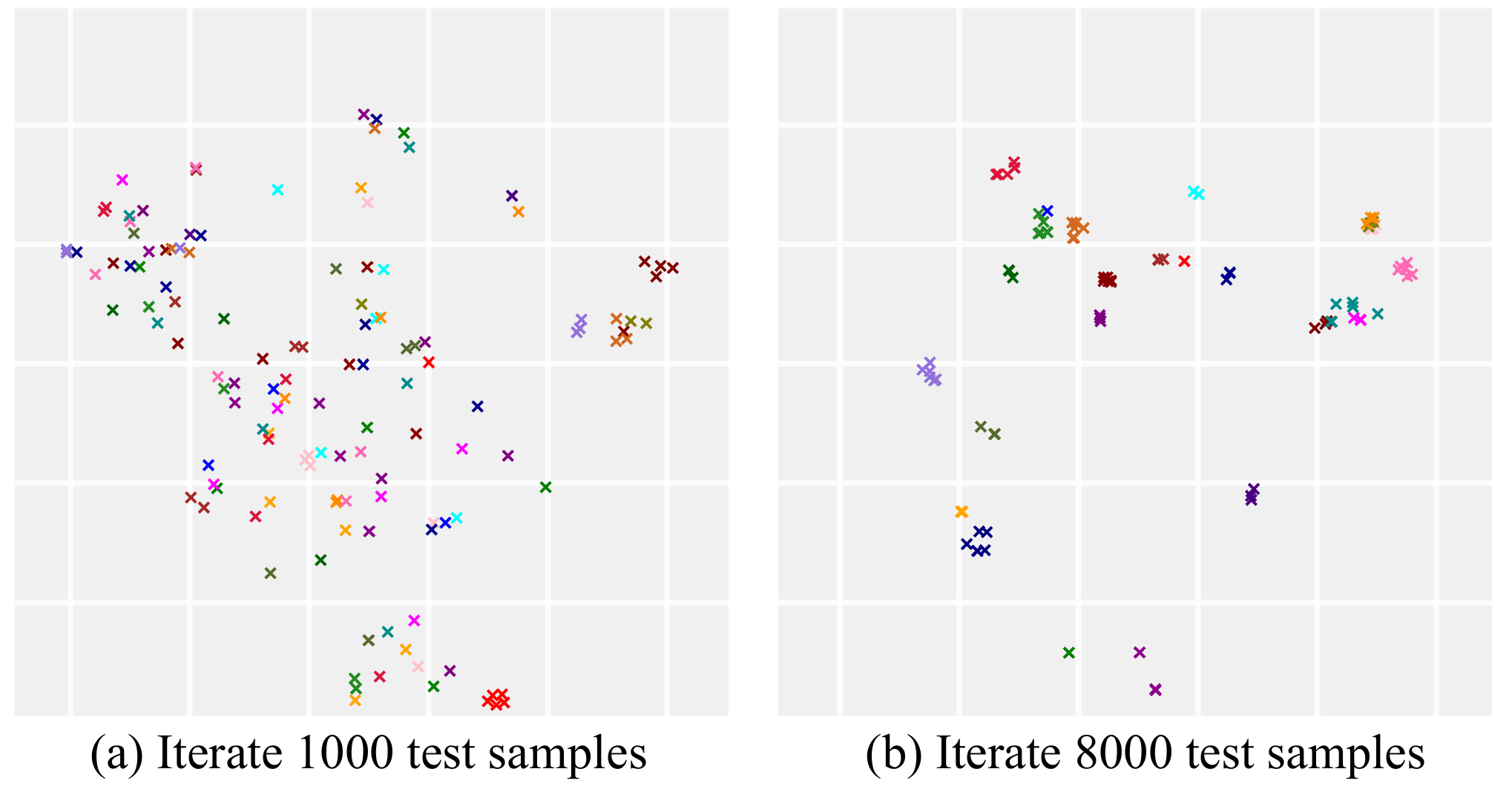}
\caption{t-SNE visualizations of cached HOI pair features from 25 classes on HICO-DET after processing 1000 test samples (left) and 8000 test samples (right), with $K_{\max}=6$.}
\label{fig:t-sne}
\end{figure}

\subsubsection{Cache Capacity and Feature Evolution}
The effect of the upper capacity bound $K_{\max}$ is illustrated in Fig.~\ref{fig:capacity}. Full and Rare mAP peak at $K_{\max}=6$ and decline at larger values, whereas Non-rare mAP varies less. Detection accuracy therefore does not increase monotonically with cache capacity. Small caches retain fewer reference features, while increasing capacity beyond the intermediate setting gives diminishing or negative returns, particularly for rare interactions.

The normalized joint selection score evolves as illustrated in Fig.~\ref{fig:quality}. The score increases rapidly during early updates and then approaches a plateau as the cache accumulates observations. Capacity 3 reaches a lower plateau, while capacity 8 provides no sustained advantage over capacity 6 in later updates. The similar late-stage scores at capacities 6 and 8 agree with the limited benefit of increasing capacity in the detection results.

Feature distributions are visualized using t-SNE~\cite{hinton2008visualizing} in Fig.~\ref{fig:t-sne}. The plots contain cached features from 25 classes after processing 1000 and 8000 test samples, with $K_{\max}=6$. Many groups become more compact as additional observations are processed, while distinct interaction groups remain visible. Shared verb semantics can also place some groups close to one another. During retrieval, the predicted object category restricts the reference entries to compatible interactions, even when classes with different objects occupy nearby regions of feature space.

\begin{figure}[!t]
\centering
\includegraphics[width=\linewidth]{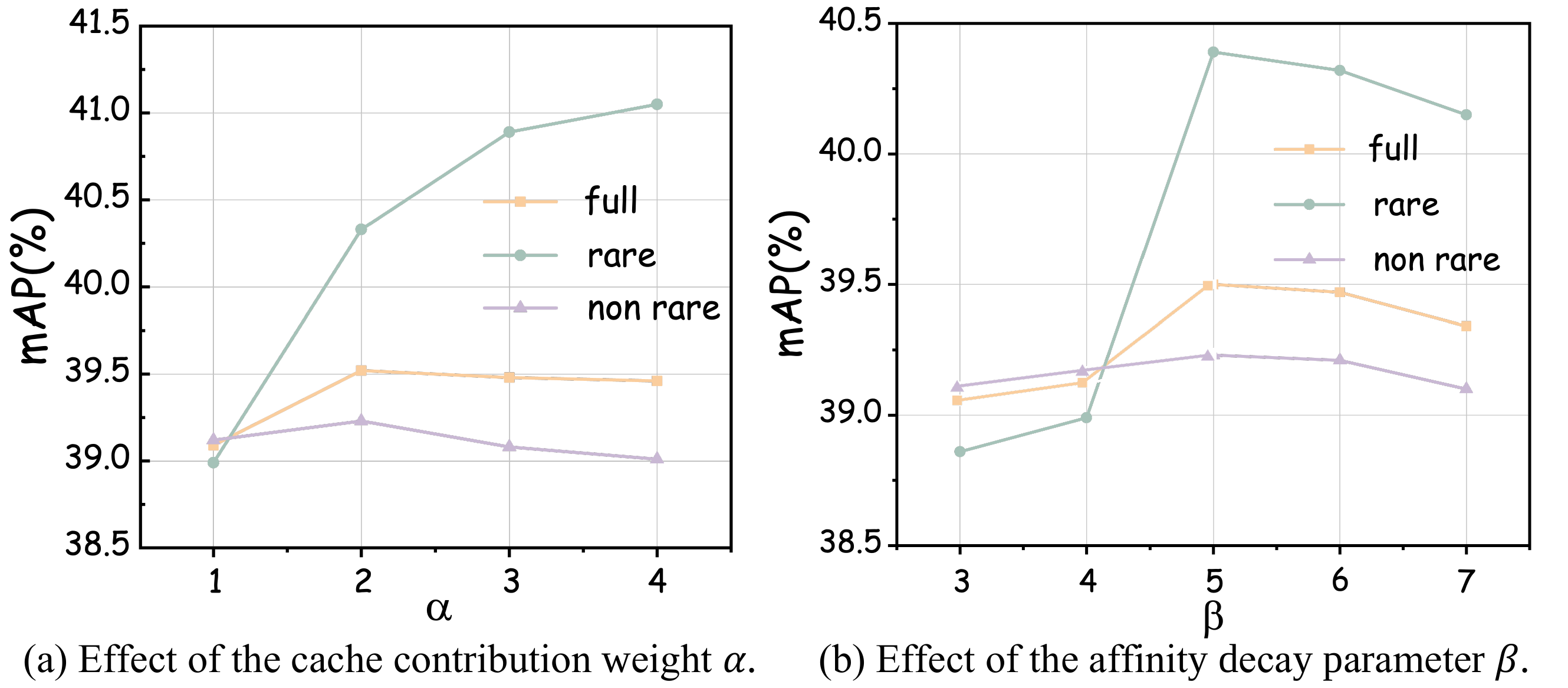}
\caption{Effect of the cache contribution $\alpha$ and affinity decay $\beta$ on detection performance. The two frequency groups respond differently to the cache weight, and the highest plotted Full mAP is obtained at $\beta=5$.}
\label{fig:alpha}
\end{figure}

\subsubsection{Hyperparameter Sensitivity}
Sensitivity to the readout parameters is evaluated in Fig.~\ref{fig:alpha}. Rare mAP responds more strongly than Full mAP to an increase in $\alpha$, whereas Non-rare mAP declines at larger cache weights. The Full curve changes relatively little near the selected value of 3. A moderate weight therefore balances the gain on rare interactions against the decline observed for frequent categories at larger values.

Full mAP peaks at $\beta=5$ and decreases for larger values in the plotted range. As $\beta$ increases, the contribution of less similar entries decays more rapidly, concentrating support around the query. A smaller value draws evidence from a broader neighborhood. The peak at an intermediate value favors a neighborhood that retains related observations without giving excessive weight to less similar entries. The settings $\alpha=3$, $\beta=5$, and $K_{\max}=6$ are used throughout the main comparisons, yielding gains across the evaluated detectors and datasets.

\begin{table}[!t]
\centering
\caption{Performance on HICO-DET under unseen-composition settings. RF-UC and NF-UC denote Rare First and Non-rare First splits, respectively.}
\label{tab:hico_results}
\small
\setlength{\tabcolsep}{3.5pt} 
\resizebox{\linewidth}{!}{%
\begin{tabular}{lcccccc}
\toprule
\multirow{2}{*}{\textbf{Method}} & \multicolumn{3}{c}{\textbf{RF-UC Setting}} & \multicolumn{3}{c}{\textbf{NF-UC Setting}} \\
\cmidrule(lr){2-4} \cmidrule(lr){5-7}
 & \textbf{Full} & \textbf{Seen} & \textbf{Unseen} & \textbf{Full} & \textbf{Seen} & \textbf{Unseen} \\
\midrule
GEN-VLKT~\cite{liao2022gen} & 30.56 & 32.91 & 21.36 & 23.71 & 23.38 & 25.05 \\
HOICLIP~\cite{ning2023hoiclip} & 32.99 & 34.85 & 25.53 & 27.75 & 28.10 & 26.39 \\
CLIP4HOI~\cite{mao2023clip4hoi} & 34.08 & 35.48 & 28.47 & 28.90 & 28.26 & 31.44 \\
LOGICHOI~\cite{li2023neural} & 33.17 & 34.93 & 25.97 & 27.95 & 27.86 & 26.84 \\
\midrule
ADA-CM~\cite{lei2023efficient} & 33.01 & 34.35 & 27.63 & 31.39 & 31.13 & 32.41 \\
\rowcolor{gray!20} \textbf{+ ADC (Ours)} & \textbf{38.17} & \textbf{38.73} & \textbf{35.96} & \textbf{36.73} & \textbf{37.17} & \textbf{34.95} \\
$\Delta$ & \textcolor{red}{+5.16} & \textcolor{red}{+4.38} & \textcolor{red}{+8.33} & \textcolor{red}{+5.34} & \textcolor{red}{+6.04} & \textcolor{red}{+2.54} \\
\midrule
HOIGEN~\cite{guo2024unseen} & 33.86 & 34.57 & 31.01 & 33.08 & 32.86 & 33.98 \\
\rowcolor{gray!20} \textbf{+ ADC (Ours)} & \textbf{35.56} & \textbf{35.78} & \textbf{34.67} & \textbf{36.01} & \textbf{36.47} & \textbf{34.19} \\
$\Delta$ & \textcolor{red}{+1.70} & \textcolor{red}{+1.21} & \textcolor{red}{+3.66} & \textcolor{red}{+2.93} & \textcolor{red}{+3.61} & \textcolor{red}{+0.21} \\
\midrule
EZ-HOI~\cite{lei2024ez} & \underline{36.73} & \underline{37.35} & \underline{34.24} & \underline{34.84} & \underline{34.47} & \underline{36.33} \\
\rowcolor{gray!20} \textbf{+ ADC (Ours)} & \textbf{37.32} & \textbf{37.39} & \textbf{37.06} & \textbf{35.44} & \textbf{34.93} & \textbf{37.48} \\
$\Delta$ & \textcolor{red}{+0.59} & \textcolor{red}{+0.04} & \textcolor{red}{+2.82} & \textcolor{red}{+0.60} & \textcolor{red}{+0.46} & \textcolor{red}{+1.15} \\
\bottomrule
\end{tabular}}
\end{table}

\subsection{Unseen-Composition Evaluation}
Unseen-composition (UC) evaluation examines interactions excluded from detector training. Rare First UC (RF-UC) prioritizes rare compositions when constructing the held-out set, whereas Non-rare First UC (NF-UC) prioritizes frequent compositions. Seen and Unseen results are reported separately in Table~\ref{tab:hico_results}. The corresponding detector training splits are retained, and ADC uses unlabeled test observations under the main two-pass protocol.

Under RF-UC, Unseen mAP improves by 8.33 points for ADA-CM, 3.66 for HOIGEN, and 2.82 for EZ-HOI. Each gain exceeds the corresponding Seen improvement of 4.38, 1.21, and 0.04 points. ADA-CM begins with the lowest Unseen score among the three detectors and obtains the largest improvement. EZ-HOI also improves on Unseen classes while its Seen score remains nearly unchanged. The larger Unseen gains indicate that the retrieved test features are especially useful for compositions absent from detector training in RF-UC.

Under NF-UC, the Unseen gains are 2.54, 0.21, and 1.15 points for ADA-CM, HOIGEN, and EZ-HOI, respectively. They are smaller than the RF-UC gains, while Seen performance also improves. In particular, HOIGEN gains 3.61 points in Seen mAP and 0.21 points in Unseen mAP. The different responses across the two splits indicate that the benefit of memory depends on the compositions held out and the initial representation of the detector. Despite their different response patterns, both splits yield positive Unseen gains for all three detectors.

\begin{table}[!t]
\centering
\caption{Performance comparison on the HICO-DET systematic generalization splits.}
\label{tab:sg_results}
\small
\setlength{\tabcolsep}{3pt}
\begin{tabular}{lcccc}
\toprule
\textbf{Method} & \textbf{Backbone} & \textbf{SG1} & \textbf{SG2} & \textbf{SG3} \\
\midrule
EZ-HOI~\cite{lei2024ez} & R50 & 16.57 & 14.62 & 15.73 \\
\rowcolor{gray!20} + ADC & R50 & \textbf{17.64} & \textbf{15.53} & \textbf{16.42} \\
$\Delta$ & - & \textcolor{red}{+1.07} & \textcolor{red}{+0.91} & \textcolor{red}{+0.69} \\
\midrule
ADA-CM~\cite{lei2023efficient} & R50 & 15.64 & 15.02 & 15.76 \\
\rowcolor{gray!20} + ADC & R50 & \textbf{17.16} & \textbf{16.93} & \textbf{17.01} \\
$\Delta$ & - & \textcolor{red}{+1.52} & \textcolor{red}{+1.91} & \textcolor{red}{+1.25} \\
\midrule
HOIGEN~\cite{guo2024unseen} & R50 & 14.40 & 13.27 & 14.53 \\
\rowcolor{gray!20} + ADC & R50 & \textbf{16.27} & \textbf{15.66} & \textbf{16.28} \\
$\Delta$ & - & \textcolor{red}{+1.87} & \textcolor{red}{+2.39} & \textcolor{red}{+1.75} \\
\bottomrule
\end{tabular}
\end{table}

\subsection{Systematic Generalization}
The HICO-DET-SG splits~\cite{takemoto2023hicosg} provide a further evaluation of generalization to novel combinations of objects and interactions. Results on SG1, SG2, and SG3 are reported in Table~\ref{tab:sg_results}, using the corresponding detector training splits and the main adaptation protocol.

The gains range from 0.69 to 1.07 points for EZ-HOI, from 1.25 to 1.91 for ADA-CM, and from 1.75 to 2.39 for HOIGEN. On SG2, ADA-CM improves from 15.02 to 16.93 and HOIGEN from 13.27 to 15.66, giving their largest gains among the three splits. All three detectors improve on every SG split, indicating that the benefit of test-time memory persists across different partitions of the interaction compositions.

\begin{table}[!t]
\centering
\caption{Detection performance, runtime, and reported GPU memory in the online setting. Time is measured over the complete test set.}
\label{tab:efficiency}
\small
\setlength{\tabcolsep}{2pt} 
\resizebox{\linewidth}{!}{
\begin{tabular}{lccccccc}
\toprule
\multirow{3}{*}{\textbf{Method}} & \multicolumn{4}{c}{\textbf{HICO-DET (Online)}} & \multicolumn{3}{c}{\textbf{V-COCO (Online)}} \\
\cmidrule(lr){2-5} \cmidrule(lr){6-8}
 & \multicolumn{2}{c}{\textbf{mAP}} & \multicolumn{2}{c}{\textbf{Cost}} & \textbf{mAP} & \multicolumn{2}{c}{\textbf{Cost}} \\
\cmidrule(lr){2-3} \cmidrule(lr){4-5} \cmidrule(lr){6-6} \cmidrule(lr){7-8}
 & \textbf{Full} & \textbf{Rare} & \textbf{Time} & \textbf{Mem.} & $\mathbf{AP^{\#1}}$ & \textbf{Time} & \textbf{Mem.} \\
\midrule
EZ-HOI~\cite{lei2024ez} & 38.33 & 36.70 & 1h 23m & 7619M & 60.50 & 23m & 6822M \\
\rowcolor{gray!10} \textbf{+ ADC} & \textbf{38.81} & \textbf{39.57} & 2h 20m & 8926M & \textbf{60.97} & 39m & 8410M \\
\midrule
ADA-CM~\cite{lei2023efficient} & 38.39 & 37.46 & 11m & 4886M & 58.50 & 8m & 4758M \\
\rowcolor{gray!10} \textbf{+ ADC} & \textbf{39.48} & \textbf{40.60} & 39m & 6908M & \textbf{61.90} & 14m & 6916M \\
\midrule
HOIGEN~\cite{guo2024unseen} & 33.56 & 32.10 & 12m & 4612M & - & - & - \\
\rowcolor{gray!10} \textbf{+ ADC} & \textbf{36.03} & \textbf{39.46} & 17m & 4920M & - & - & - \\
\bottomrule
\end{tabular}
}
\end{table}

\subsection{Efficiency and Online Adaptation}
\label{sec:adc_efficiency}
\subsubsection{Runtime and Memory}
Runtime and GPU memory in the online setting are reported in Table~\ref{tab:efficiency}. On HICO-DET, inference time increases from 83 to 140 minutes for EZ-HOI, from 11 to 39 minutes for ADA-CM, and from 12 to 17 minutes for HOIGEN. The corresponding runtime ratios are $1.69\times$, $3.55\times$, and $1.42\times$. Reported GPU memory increases by 17.2\%, 41.4\%, and 6.7\%, respectively. The additional cost comes from cache updates, retrieval, and augmentation, with all model parameters kept fixed.

The accuracy--cost tradeoff varies across detectors. HOIGEN obtains a 7.36-point Rare gain with five additional minutes and 308M more reported GPU memory. ADA-CM obtains a 3.14-point Rare gain with a larger relative increase in runtime. On V-COCO, inference time changes from 23 to 39 minutes for EZ-HOI and from 8 to 14 minutes for ADA-CM, with GPU memory increasing by approximately 23.3\% and 45.4\%. The additional inference cost is therefore detector dependent, with the smallest relative overhead observed for HOIGEN.

\subsubsection{Online Detection Performance}
With an initially empty cache, ADC improves HICO-DET Full mAP by 0.48, 1.09, and 2.47 points for EZ-HOI, ADA-CM, and HOIGEN, respectively. Rare mAP increases by 2.87, 3.14, and 7.36 points. The positive Rare gains across all three detectors show that a populated initial cache is not required to improve underrepresented interactions.

The online setting retains approximately 56\%, 78\%, and 86\% of the corresponding two-pass Rare gains for EZ-HOI, ADA-CM, and HOIGEN. Full mAP is lower than the two-pass result by 0.99, 0.33, and 1.92 points, respectively. On V-COCO, online role mAP improves by 0.47 points for EZ-HOI and 3.40 points for ADA-CM. The two-pass protocol yields larger gains, while single-pass adaptation retains substantial improvements as the cache accumulates entries.

\begin{figure}[!t]
\centering
\includegraphics[width=\linewidth]{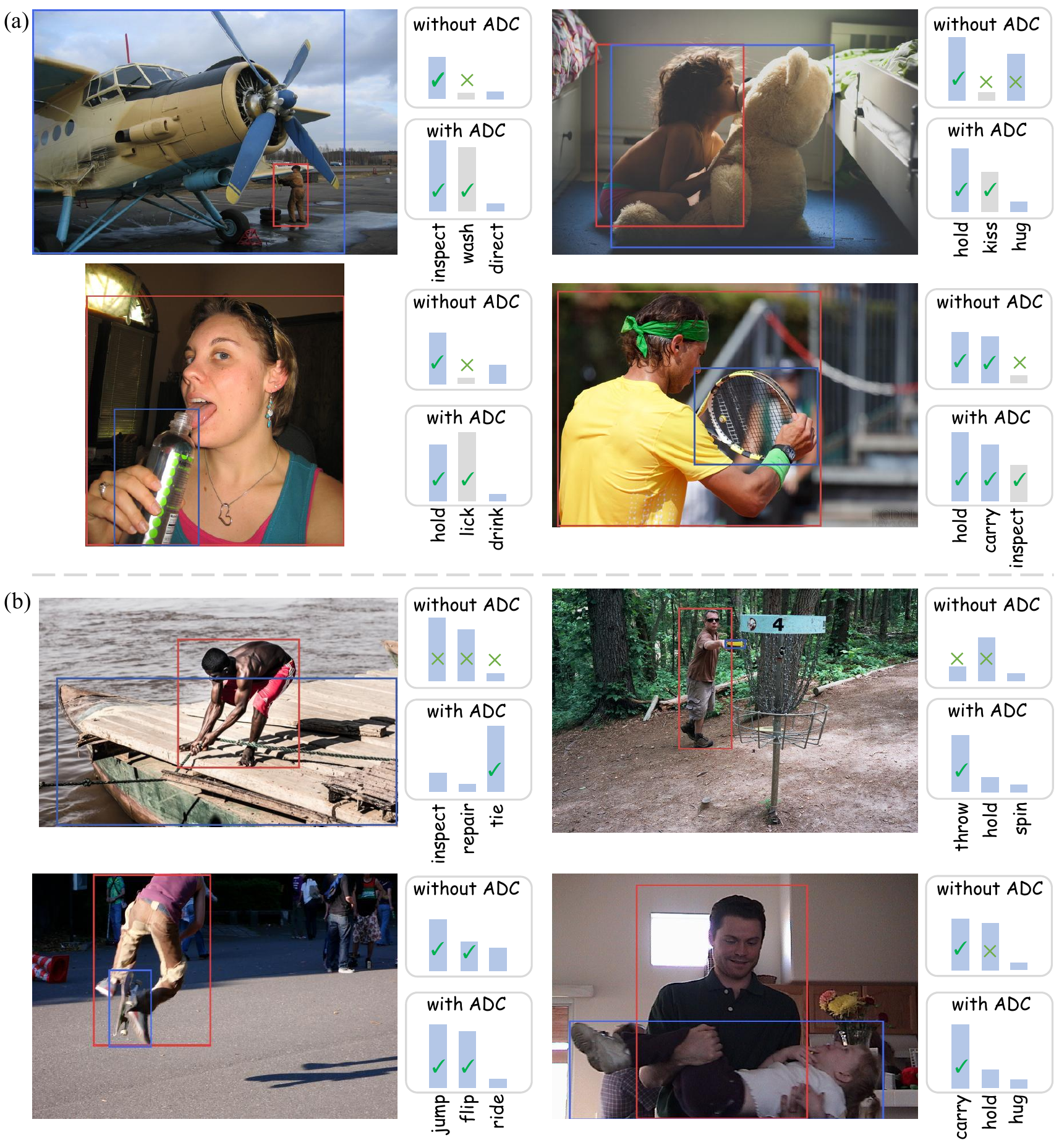}
\caption{Qualitative comparison of ADA-CM with and without ADC. Gray bars indicate rare-category predictions, and blue bars indicate non-rare-category predictions. (a) Examples with rare interactions. (b) Examples without rare interactions.}
\label{fig:qualitative}
\end{figure}

\subsection{Qualitative Results}
Qualitative comparisons based on ADA-CM are illustrated in Fig.~\ref{fig:qualitative}. In panel (a), the retrieved entries raise the scores of rare interactions that are missed or weakly scored by the base detector, helping recover the missing detections.

Panel (b) illustrates the effect on several plausible interactions associated with the same human--object pair. By retrieving object-compatible features, ADC adjusts their relative support, favoring the annotated interactions among competing candidates while retaining multiple valid labels.

\section{Conclusion}
We propose ADC to improve long-tailed HOI detection through training-free test-time memory. FACA distributes bounded capacity according to training class frequencies and supplements underfilled caches with selected augmented views. CJCS retains entries by combining prediction confidence with local feature diversity. A constrained memory formulation connects the two decisions, with the allocation and ranking rules characterized through weighted logarithmic utility and constrained selection. Experiments on HICO-DET and V-COCO yield consistent gains across three detectors, with the largest improvements on rare interactions. ADC also improves recognition of unseen compositions and remains effective with caches built incrementally during inference. ADC thus offers a plug-and-play addition to trained HOI detectors while keeping their network parameters fixed.

\appendices
\section{Capacity Allocation and Rounding}
\label{app:allocation}
The Hessian of $U$ in Eq.~\eqref{eq:utility} is $-\operatorname{diag}(r_c/k_c^2)$ and is negative definite. The objective is therefore strictly concave on the feasible box. For $CK_{\min}<B<CK_{\max}$, it is increasing in every coordinate, so the budget constraint is active. The Lagrangian is
\begin{equation}
\begin{aligned}
\mathcal L={}&\sum_c r_c\log k_c-\mu\left(\sum_c k_c-B\right)\\
&+\sum_c\xi_c(k_c-K_{\min})+\sum_c\zeta_c(K_{\max}-k_c),
\end{aligned}
\label{eq:lagrangian}
\end{equation}
where $\mu$, $\xi_c$, and $\zeta_c$ are nonnegative multipliers. Stationarity gives $r_c/k_c-\mu+\xi_c-\zeta_c=0$. At an interior coordinate, complementary slackness yields $k_c=r_c/\mu$. At the lower bound, $r_c/\mu\leq K_{\min}$, and at the upper bound, $r_c/\mu\geq K_{\max}$. These conditions give Eq.~\eqref{eq:waterfill}. Strict concavity ensures a unique capacity vector. At $B=CK_{\min}$, every class receives $K_{\min}$, and for $B\geq CK_{\max}$, every class receives $K_{\max}$.

For a fixed scale, let $\bar k_c=\operatorname{Clip}(K_{\mathrm{base}}r_c,K_{\min},K_{\max})$ and $B_0=\sum_c\bar k_c$. This allocation satisfies the same optimality conditions at its induced budget with $\mu=1/K_{\mathrm{base}}$. With integer bounds, Eq.~\eqref{eq:capacity} gives $K_c=\lfloor\bar k_c\rfloor$, so the difference in total capacity satisfies
\begin{equation}
0\leq B_0-\sum_c K_c<C.
\label{eq:roundslots}
\end{equation}
The loss in the logarithmic allocation utility is bounded by
\begin{equation}
0\leq U(\bar{\boldsymbol k})-U(\boldsymbol K)
\leq\sum_c r_c\frac{\bar k_c-K_c}{K_c},
\label{eq:roundutility}
\end{equation}
which follows by applying $\log(1+x)\leq x$ to each coordinate. Thus, integer rounding preserves feasibility and controls the loss in the allocation objective.

When the bounds are inactive, $k_c^*=Br_c/R$, where $R=\sum_c r_c$. Relative to a feasible uniform allocation $k_c^{\mathrm{unif}}=B/C$, the utility difference is
\begin{equation}
U(\boldsymbol k^*)-U(\boldsymbol k^{\mathrm{unif}})
=R\sum_c\bar r_c\log(C\bar r_c)\geq0,
\label{eq:utilitygain}
\end{equation}
where $\bar r_c=r_c/R$. The right-hand side is $R$ times the relative entropy between the normalized priority distribution and the uniform distribution. It is zero for equal priorities and positive when priorities differ, expressing the advantage of adapting capacity to class weights under the stated utility.

\section{Selection and Local Feature Variation}
\label{app:selection}
For a fixed pool, the Lagrangian relaxation of Eq.~\eqref{eq:selection_constraint} is
\begin{equation}
D(\mathcal S)+\nu[Q(\mathcal S)-q_0],\qquad \nu\geq0.
\label{eq:selection_lagrangian}
\end{equation}
Removing the constant and dividing by $1+\nu$ yields Eq.~\eqref{eq:joint}. If a selected entry has a lower joint score than an unselected one, exchanging them increases the objective. Repeating this operation gives the top-$m_c$ set and proves the first statement of Proposition~\ref{prop:selection}. For $0<\tau<1$, any equally sized set that improves either $D$ or $Q$ without reducing the other has a larger weighted sum. Optimality of the weighted sum therefore gives Pareto optimality for the pool-based objectives. The scores used in this argument are those evaluated on the fixed temporary pool, and $\tau$ specifies the tradeoff in the relaxed objective.

The local diversity contribution also has a direct geometric interpretation. For unit-normalized features at distance $r\in[0,2]$, $\|f_i-f_j\|_2^2=2(1-f_i^\top f_j)$, giving
\begin{equation}
\varphi(r)=\tfrac12r^2e^{-\gamma r^2},\qquad
\varphi'(r)=re^{-\gamma r^2}(1-\gamma r^2).
\label{eq:kernel}
\end{equation}
The maximum is attained at $\min\{2,\gamma^{-1/2}\}$. For $\gamma>1/4$, the contribution first increases and then decreases within the admissible range. Nearly identical features and sufficiently distant features consequently contribute less than those at an intermediate distance. For $\gamma\leq1/4$, the contribution increases up to the endpoint $r=2$. This calculation characterizes the interaction between angular variation and locality under unit normalization. For general nonzero features, Eq.~\eqref{eq:diversity} retains angular difference and Euclidean proximity as separate factors.

\section{Retrieval Sensitivity}
\label{app:stability}
Consider fixed retrieved keys and compare the cache values with reference values for the same pairs. For an incorrectly assigned entry, let $v_i^*$ be a one-hot vector of a valid interaction, or zero when the pair has no valid interaction. The score difference can be written as
\begin{equation}
z(q)-z^*(q)=\alpha\sum_{i\in\mathcal E}w_i(q)(e_{\ell_i}-v_i^*),
\label{eq:valueerror}
\end{equation}
where $\mathcal E$ denotes the entries whose values differ and $z^*$ uses the same base prediction and keys with reference values. Since $\|e_{\ell_i}-v_i^*\|_\infty\leq1$, the triangle inequality gives
\begin{equation}
\|z(q)-z^*(q)\|_\infty
\leq\alpha\sum_{i\in\mathcal E}w_i(q)=:\epsilon(q).
\label{eq:noise}
\end{equation}
Thus, the influence of the changed assignments is controlled by their total affinity weight. In particular, a reference class-score margin larger than $2\epsilon(q)$ is preserved because
\begin{equation}
z_a-z_b\geq z_a^*-z_b^*-2\epsilon(q)>0.
\label{eq:margin}
\end{equation}
If the affected entries satisfy $q^\top\hat f_i\leq\rho<1$, their total influence is bounded by $\alpha|\mathcal E|e^{-\beta(1-\rho)}$. This expresses the attenuation of support from distant entries in the retrieval rule.

For augmentation, compare a normalized augmented key $\hat f_j^+$ with a normalized parent copy $\hat f_j^0$ in the same slot and with the same class value. If $\|\hat f_j^+-\hat f_j^0\|_2\leq\varepsilon_j$, their inner products with a unit query differ by at most $\varepsilon_j$. The derivative of $e^{\beta(a-1)}$ is at most $\beta$ on $a\in[-1,1]$. Applying the mean-value theorem to $m$ such entries yields
\begin{equation}
\|z_{\mathrm{aug}}-z_{\mathrm{parent}}\|_\infty
\leq\alpha\beta\sum_{j=1}^{m}\varepsilon_j.
\label{eq:augerror}
\end{equation}
This bound quantifies the effect of feature changes at fixed occupancy. Adding $m$ entries contributes at most $\alpha m$ in 1-norm because each retrieval weight is at most one. The two expressions distinguish changes in the appearance of stored features from the increase in support obtained by occupying additional slots.

\section{Computational Complexity}
\label{app:complexity}
Let $B$ bound the total number of persistent cache entries. Storing $d$-dimensional features and $C$-dimensional prediction vectors requires $O(B(d+C))$ scalars. For a query with $M_o$ object-compatible entries, computing the affinities costs $O(M_od)$. With class values stored as indices, accumulating their scores costs $O(M_o+C)$.

For a candidate pool of size $M_c$, direct evaluation of the diversity score costs $O(M_c^2d)$, and entropy computation costs $O(M_cC)$. Selecting the highest-scoring entries can be implemented by sorting, with cost $O(M_c\log M_c)$. A single-entry update has $M_c\leq K_c+1$, so the class capacity directly controls the size of this computation. Completion requires forward passes for augmented inputs and a temporary candidate pool. The persistent memory remains bounded by the allocated capacities, and all operations are performed without backpropagation through the detector.

\bibliographystyle{IEEEtran}
\bibliography{reference}
\end{document}